\documentclass{article}
\usepackage{graphicx}
\usepackage{amsmath}
\usepackage{hyperref}
\usepackage{tabularx}
\usepackage{multirow}
\usepackage{float}
\usepackage{booktabs}

\title{DMCD: Semantic-Statistical Framework for Causal Discovery}
\author{
  Samarth KaPatel, 
  Sofia Nikiforova\thanks{Corresponding author.},\\
  Giacinto Paolo Saggese,
  Paul Smith \\
  Causify AI \\
  \texttt{\{s.kapatel, s.nikiforova, gp, paul\}@causify.ai} \\
}
\date{}

\begin{document}

\maketitle

\begin{abstract}
We present \textbf{DMCD (DataMap Causal Discovery)}, a two-phase causal discovery framework that integrates LLM-based semantic drafting from variable metadata with statistical validation on observational data. In Phase I, a large language model proposes a sparse draft DAG, serving as a semantically informed prior over the space of possible causal structures. In Phase II, this draft is audited and refined via conditional independence testing, with detected discrepancies guiding targeted edge revisions.

We evaluate our approach on three metadata-rich real-world benchmarks spanning industrial engineering, environmental monitoring, and IT systems analysis. Across these datasets, DMCD achieves competitive or leading performance against diverse causal discovery baselines, with particularly large gains in recall and F1 score. Probing and ablation experiments suggest that these improvements arise from semantic reasoning over metadata rather than memorization of benchmark graphs. Overall, our results demonstrate that combining semantic priors with principled statistical verification yields a high-performing and practically effective approach to causal structure learning.
\end{abstract}

\section{Introduction}
\label{sec:intro}

Causal discovery from observational data is central to scientific inquiry, enabling reasoning about interventions, counterfactuals, and the underlying structure of complex phenomena. Formally, the objective is to recover a directed acyclic graph (DAG) that encodes structural relationships among observed variables.

Inferring such structure from observational data alone is fundamentally challenging. The space of possible DAGs grows super-exponentially with the number of variables, forcing causal discovery algorithms to navigate a vast combinatorial landscape using statistical tests or optimization procedures. In effect, these methods attempt to constrain an enormous search space using only patterns present in the data. 

Yet in many real-world settings, additional information is available. Variable names, descriptions, and other metadata often encode meaningful domain knowledge that can help narrow the space of plausible structures. Classical causal discovery approaches, however, typically abstract away from this information. While this abstraction may be appropriate in curated synthetic benchmarks designed to isolate statistical performance, it is less natural in practical applications --- particularly if the goal is to produce structures that accurately reflect domain mechanisms and support practitioner decision-making.

In practice, human experts routinely rely on semantic reasoning when forming causal hypotheses, using domain knowledge to restrict the space of plausible graphs before turning to data for validation. Recent advances in large language models (LLMs) provide a new opportunity to operationalize this behavior: LLMs can interpret variable metadata and draw on broad background knowledge to act as proxies for human experts during causal discovery. In light of this, we introduce \textbf{DMCD (DataMap Causal Discovery)}, a hybrid framework that integrates LLM-based semantic reasoning with empirical verification. DMCD is structured as a two-phase pipeline. In Phase I, an LLM interprets variable metadata to generate a draft causal graph, effectively serving as a semantic prior over DAG space. In Phase II, this draft is evaluated and refined through conditional independence testing on observational data.

DMCD constitutes the causal discovery component of a broader system, Causify DataMap, which serves as a causal knowledge layer for domain practitioners~\cite{causify2026datamap}. Beyond discovery, DataMap includes functionality for data preprocessing, interactive graph refinement and explanation, and downstream causal analysis, such as interventional and counterfactual reasoning. In this paper, however, we focus exclusively on the causal discovery pipeline. 

We evaluate DMCD on three real-world causal discovery benchmarks spanning industrial engineering, environmental monitoring, and IT systems analysis, each preserving meaningful variable metadata. Across datasets, DMCD achieves competitive or leading performance relative to established constraint-based, score-based, functional, and continuous optimization methods, with particularly large gains in recall and F1 score. Our results demonstrate that incorporating semantic priors derived from variable metadata can substantially strengthen data-driven causal discovery, suggesting a promising direction for hybrid semantic-statistical structure learning.

\section{Related work}
\label{sec:related_work}

Causal discovery has traditionally been approached primarily through data-driven statistical methods. Recent advances in LLMs, however, have opened new avenues for incorporating semantic information into a variety of causal tasks, including structure learning~\cite{wan2025large, ma2025causal}. We situate DMCD within this emerging landscape. 

\paragraph{LLMs for Independent Causal Discovery}
A growing body of work employs LLMs as self-sufficient agents for causal discovery~\cite{willig2022can, kiciman2023causal, long2023can, jiralerspong2024efficient,newsham2025large,vashishtha2025causal}. In this setting, LLMs are provided with textual descriptions of variables and asked to infer pairwise or global causal relations. Such approaches treat LLMs as domain experts capable of extracting causal structure from semantic information alone.

While these methods demonstrate that LLMs possess non-trivial causal reasoning capabilities, they typically do not incorporate systematic statistical validation against observational data. As a result, their outputs may reflect semantic plausibility without empirical grounding. In contrast, DMCD combines metadata-based drafting with targeted statistical verification, ensuring that the final structure is evaluated against data-derived evidence. 

\paragraph{LLMs for Posterior Correction}
Another line of research uses LLMs to refine causal structures produced by traditional statistical algorithms~\cite{long2023causal, ban2023causal, takayama2024integrating,khatibi2024alcm}. In these frameworks, a partially identified graph --- often representing a Markov equivalence class --- is presented to an LLM, which then orients edges or reduces the equivalence class using contextual reasoning and background knowledge.

DMCD differs by reversing the typical pipeline in this class of approaches, in which statistical discovery is followed by semantic adjustment. In DMCD, the initial graph is generated through semantic reasoning over metadata, and statistical validation operates as a downstream evaluation and refinement mechanism.

\paragraph{LLMs for Prior Knowledge}
A third paradigm incorporates LLM-derived knowledge as prior information within established causal discovery algorithms~\cite{ban2025integrating,shen2025exploring,kampani2024llm,darvariu2024large,li2024realtcd}. Prior knowledge may be introduced as hard constraints that eliminate incompatible edges, or as soft probabilistic regularizers integrated into score-based methods (e.g., by augmenting scoring functions with edge priors) or continuous optimization frameworks (e.g., by initializing adjacency matrices).

Compared to these approaches, DMCD does not encode LLM outputs as constraints within a statistical objective or scoring framework. Instead, it treats the LLM as a primary hypothesis generator that proposes an explicit draft DAG, which is subsequently audited and selectively revised using statistics-based discrepancy signals. This preserves transparency in how semantic hypotheses interact with empirical evidence, rather than embedding LLM-derived priors implicitly within an optimization procedure.

\paragraph{Summary}
Existing work leverages LLMs either to directly produce causal structure, to refine statistically learned graphs, or to inject prior constraints into traditional algorithms. DMCD instead formalizes the interaction between semantic reasoning and statistical evidence as an explicit hypothesis--validation pipeline: the LLM generates a draft DAG, which is then subjected to systematic empirical testing. This design preserves interpretability while enabling principled integration of knowledge-based and data-based inference.

\section{DMCD Architecture}
\label{sec:dmcd_architecture}

\begin{figure}
\centerline{\includegraphics[width=\linewidth]{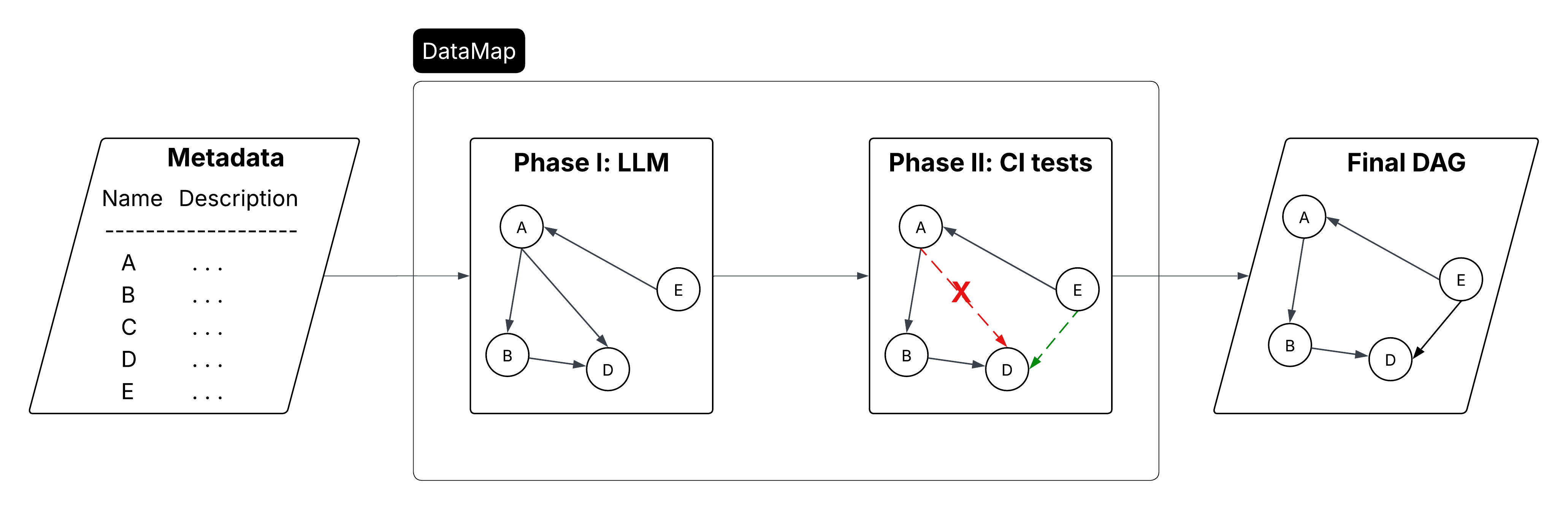}}
\caption{Overview of DMCD architecture.}
\label{fig:dmcd_architecture}
\end{figure}

The DMCD causal discovery algorithm follows a two-phase pipeline illustrated in Figure~\ref{fig:dmcd_architecture}. Phase I generates a draft causal graph using an LLM by interpreting variable metadata as semantic cues for plausible causal relations. Phase II performs statistical verification through conditional independence testing, outputting a final causal graph.

\subsection{Phase I: LLM-based drafting}
\label{sec:phase_i_llm_based_drafting}

Given a dataset with variables \(V = \{X_1, \dots, X_n\}\) and associated metadata \(M = \{m_1, \dots, m_n\}\) (e.g., variable names, descriptions, units), DMCD constructs an LLM prompt that serializes \(M\) into a structured textual representation and appends a set of high-level drafting instructions. These instructions 
\begin{itemize}
    \item position the LLM as an expert in the automatically inferred domain of the dataset;
    \item impose structural constraints requiring the output to be a directed acyclic graph, prohibiting isolated nodes, and enforcing that node labels exactly match the dataset's variable identifiers;
    \item provide guidance to include all plausible causal relationships among variables that represent stable, generalizable factors, while carefully determining the most appropriate causal direction for each edge.
\end{itemize}
The LLM then processes this prompt, utilizing information provided by \(M\) as well as relevant background and contextual knowledge, and returns its proposal in a fixed structured format. The output specifies a subset of variables \(V_{\text{draft}} \subseteq V\) deemed to participate in causal relations, together with a directed edge set \(E_{\text{draft}} \subseteq V_{\text{draft}} \times V_{\text{draft}}\), thereby defining the draft DAG \(G_{\text{draft}} = (V_{\text{draft}}, E_{\text{draft}})\).

A series of checks is performed on the proposed DAG, to ensure it is fully supported by data and well-formed. In particular, every node in \( G_{\text{draft}}\) is required to correspond to a variable in \(V\), preventing the introduction of variables with no observational support. If the LLM identifies variables that are potentially causally relevant but are missing from the dataset, DMCD reports them separately as candidates for expansion of data coverage.

This phase aims to emulate the behavior of a domain expert who first proposes a tentative causal graph before turning to data for validation. Expert-curated graphs are typically sparse: many measured variables are excluded as conceptually irrelevant, and only a limited subset of possible edges is considered plausible. Phase I of DMCD mirrors this practice by selecting a subset of variables \(V_{\text{draft}}\) and proposing a sparse draft structure over them. At this stage, DMCD, consistent with expert workflows, relies on semantic and background knowledge rather than statistical evidence. The subsequent phase provides principled empirical validation of the proposed structure.

\subsection{Phase II: Verification with data} 
\label{sec:phase_ii_verification_with_data}

Given the draft structure \( G_{\text{draft}}\), DMCD evaluates whether the causal assumptions implied by the graph are consistent with the available observational data. Rather than enforcing strict alignment, Phase II treats discrepancies between the draft graph and statistical evidence as signals for refinement.

Specifically, DMCD enumerates all unordered pairs of variables in \(V_{\text{draft}}\) and compares the conditional independence (CI) relations implied by \( G_{\text{draft}}\) to those estimated from data. A $p$-value is computed for each variable pair using an appropriate CI test (see Section~\ref{sec:implementation}). Since typically a large number of CI tests are performed, raw $p$-values are adjusted to $q$-values using a false discovery rate control method~\cite{storey2003statistical}, to mitigate the risk of false positive conclusions due to multiple hypothesis testing. Importantly, CI testing is performed exclusively using statistical procedures rather than the LLM. This ensures that empirical validation is driven by data alone and is not directly influenced by stochastic variation in LLM outputs.

Two types of discrepancies may be identified:
\begin{itemize}
    \item Potentially missing edges: pairs of variables that appear statistically dependent but are $d$-separated in the graph.
    \item Potentially spurious edges: pairs of variables that are connected in the graph while the data provides weak or inconsistent evidence of dependence. 
\end{itemize}

If discrepancies are detected, they are presented to the LLM for structural revision under predefined consistency guidelines that balance statistical strength and semantic plausibility. The graph may therefore be updated to incorporate justified edge additions or removals, but certain discrepancies may also be retained if statistical evidence is weak or unstable. For example, two variables can be marginally independent yet become conditionally dependent after conditioning on a common effect (collider bias). In such cases, the induced dependence does not justify a direct causal edge between the pair, and DMCD therefore refrains from adding one.

The resulting structure defines the final DAG:
\[
G_{\text{final}} = (V_{\text{draft}}, E_{\text{final}}),
\]
where \(E_{\text{final}} \subseteq V_{\text{draft}} \times V_{\text{draft}}\).

\subsection{Implementation}
\label{sec:implementation}

We implement the drafting phase using GPT-5.2 via the \href{https://openai.com/api/}{OpenAI API}. The temperature is set to $0.0$ to decrease variability in the outputs; we note, however, that determinism cannot be guaranteed even under ``deterministic'' decoding settings due to implementation-level stochasticity --- an issue that we discuss further in Section~\ref{sec:non_determinism}.

The verification phase utilizes conditional independence tests from the \texttt{pgmpy} Python package~\cite{ankan2024}. For continuous data, we employ a linear partial correlation-based test (implemented via Pearson correlation on regression residuals). For discrete data, we use the Chi-squared test. For mixed data, we apply a test that regresses $X$ and $Y$ on $Z$ using an XGBoost estimator and then performs an association test on the residuals using a Pillai's Trace statistic~\cite{ankan2023simple}. We use significance level $\alpha = 0.05$ throughout. 

\section{Datasets}
\label{sec:datasets}

A fundamental challenge in evaluating causal discovery methods such as DMCD is the scarcity of real-world benchmarks with meaningful metadata. Many causal discovery datasets~\cite{geffner2022deep, cheng2023causaltime, munoz2020causeme} anonymize variable names (e.g., using labels like ``X1'', ``X2'') to focus evaluation purely on algorithmic pattern recognition, deliberately abstracting away domain semantics. While valuable for testing statistics-based approaches, such datasets are unsuitable for methods like DMCD that leverage domain knowledge through variable metadata.

To evaluate DMCD under realistic, metadata-rich conditions, we select three publicly available causal discovery benchmarks that preserve meaningful variable semantics and represent diverse real-world domains: industrial processes, environmental science, and IT systems monitoring.

\subsection{Tennessee Eastman Process dataset}
\label{sec:tennessee_eastman_process_dataset}

The Tennessee Eastman Process is a benchmark chemical process simulation~\cite{downs1993plant} that is frequently used to evaluate models for various tasks, such as anomaly detection, predictive monitoring, and process control \cite{reinartz2021extended, bathelt2015revision, ricker1995nonlinear}. We use the version proposed for benchmarking causal discovery methods in~\cite{menegozzo2022cipcad}, which provides 33 variables with original engineering tags (e.g., ``XMEAS(9)'') as well as descriptions (e.g., ``Reactor Temperature'') and units (``Deg C''). The true causal structure is derived from the description of the plant in~\cite{chen2020complex} and reported in~\cite{menegozzo2022cipcad}.

\subsection{Fluxnet2015 dataset}
\label{sec:fluxnet2015_dataset}

The Fluxnet2015 dataset~\cite{pastorello2020fluxnet2015} provides standardized measurements from global meteorological stations. We use the subset processed for causal discovery by~\cite{melkas2021interactive}, which includes six meaningful variables, such as \emph{shortwave downward radiation} and \emph{vapour pressure deficit}, each accompanied by a brief description. In lieu of a ground truth DAG, \cite{melkas2021interactive} provide a DAG constructed by a domain expert drawing from a portion of the available data.

\subsection{IT monitoring datasets}
\label{sec:it_monitoring_datasets}

Four IT monitoring datasets proposed in~\cite{ait2023case} contain data collected using \href{https://www.nagios.org/}{Nagios}, an open-source software that monitors systems, networks and infrastructure. The datasets cover Middleware oriented Message activity (seven variables like \emph{percentage of used CPU}, \emph{number of consumers}), message ingestion activity (eight variables like \emph{extraction of metrics from messages}, \emph{insertion of data in a database}), web server activity (ten variables like \emph{number of PHP processes}, \emph{percentage of RAM used by all HTTP processes}), and antivirus activity (thirteen variables like \emph{percentage of memory usage of antivirus process}, \emph{Disk IO read in Kbytes/second}). Ground truth causal graphs provided in the paper were constructed either by IT monitoring experts or directly derived from system topology. 

\section{Results}
\label{sec:results}

Our evaluation follows a paper-to-paper comparison approach: we apply DMCD to the datasets described in Section~\ref{sec:datasets} and compare its performance to the results reported in the original publications. For each dataset, we compare against the baseline algorithms reported in those papers and compute the same evaluation metrics (using original implementations when available).

This approach ensures fair comparison while respecting each benchmark's established evaluation protocol. Since we do not re-implement or re-run competitor algorithms, we avoid potential implementation discrepancies and directly compare against published results.

\paragraph{Metrics} Whenever possible, we adopt the metrics used in each original paper. For the Tennessee Eastman Process dataset, we implement the following metrics in strict accordance with the equations given in the paper: False discovery rate (FDR), True positive rate (TPR), False positive rate (FPR), Structural Hamming Distance (SHD), Precision, Recall, and F1 score. For the Fluxnet2015 dataset, where the original paper provides the DAGs produced by other algorithms but no quantitative metrics, we apply the same evaluation as for the Tennessee Eastman Process dataset, with one modification: we bound FDR, FPR, Precision, and F1 score to [0, 1] for consistency. For the IT monitoring datasets, we use their provided code to compute F1 score, which is the only metric reported in the paper. 

\paragraph{Reported DMCD values} Due to inherent variability in LLM responses, DMCD may produce slightly different graphs across runs. We address this by executing DMCD 10 times on each dataset with identical hyperparameters (see Section~\ref{sec:implementation}) and reporting both mean metric values and standard deviations.

\begin{table}[h]
\centering
\begin{tabular}{l l l l r}
\toprule
& \textbf{FDR} $\downarrow$ & \textbf{TPR} $\uparrow$ & \textbf{FPR} $\downarrow$ & \textbf{SHD} $\downarrow$ \\
\midrule
CORL & 0.7812 & 0.0427 & 0.0687 & 176 \\
DirectLiNGAM & 0.9062 & 0.0455 & 0.0628 & 89 \\
FCI & 0.9375 & 0.1667 & 0.0581 & 37 \\
GES & 0.875 & 0.198 & 0.0656 & 106 \\
GOLEM & 0.9062 & 0.0306 & 0.0674 & 120 \\
ICALiNGAM & 0.7812 & 0.0787 & 0.0569 & 100 \\
MCSL & 0.9375 & 0.08 & 0.0596 & 53 \\
NOTEARS & 1 & 0 & 0.0612 & \textbf{35} \\
NOTEARS-MLP & 0.9375 & 0.0364 & 0.0634 & 82 \\
PC & \textbf{0.75} & 0.1695 & 0.0512 & 69 \\
\hline
\textbf{DMCD} & $1.199 \pm 0.05$ & $\mathbf{0.236 \pm 0.04}$ & $\mathbf{0.033 \pm 0.004}$ & $44.3 \pm 3.52$ \\
\bottomrule
\end{tabular}
\vspace{0.5cm}
\begin{tabular}{l l l l}
\toprule
& \textbf{Precision} $\uparrow$ & \textbf{Recall} $\uparrow$ & \textbf{F1} $\uparrow$ \\
\midrule
CORL & 0.2188 & 0.0427 & 0.0714 \\
DirectLiNGAM & 0.0938 & 0.0455 & 0.0612 \\
FCI & 0.0625 & 0.1667 & 0.0909 \\
GES & 0.125 & 0.0396 & 0.0602 \\
GOLEM & 0.0938 & 0.0306 & 0.0462 \\
ICALiNGAM & 0.2188 & 0.0787 & 0.1157 \\
MCSL & 0.0625 & 0.08 & 0.0702 \\
NOTEARS & 0 & 0 & 0 \\
NOTEARS-MLP & 0.0625 & 0.0364 & 0.046 \\
PC & \textbf{0.25} & 0.1356 & 0.1758 \\
\hline
\textbf{DMCD} & $0.189 \pm 0.04$ & $\mathbf{0.236 \pm 0.04}$ & $\mathbf{0.209 \pm 0.04}$ \\
\bottomrule
\end{tabular}
\caption{Performance comparison on Tennessee Eastman Process dataset. Best scores for each metric are highlighted in \textbf{bold}.}
\label{tab:tep_results}
\end{table}

Table~\ref{tab:tep_results} demonstrates the results of applying DMCD to the Tennessee Eastman Process benchmark alongside ten established causal discovery methods. While NOTEARS achieves the lowest SHD and PC obtains the best FDR and Precision, DMCD outperforms all competitors in TPR, FPR, Recall, and F1 score. These results indicate a favorable balance between discovery capability and false positive control, which is particularly relevant in industrial monitoring settings. The FDR exceeding 1 stems from the benchmark's specific metric definition that does not normalize by reversed edges. When bounded to [0, 1], DMCD achieves an FDR of $0.869 \pm 0.03$, which is on par with the other algorithms.

\begin{table}[h]
\centering
\begin{tabular}{l l l l r}
\toprule
 & \textbf{FDR} $\downarrow$ & \textbf{TPR} $\uparrow$ & \textbf{FPR} $\downarrow$ & \textbf{SHD} $\downarrow$ \\
\midrule
PC Pearson & 0.5556 & 0.5714 & 0.2381 & 8 \\
PC Spearman & 0.7143 & 0.2857 & 0.2381 & 10 \\
LiNGAM & 0.6667 & 0.2857 & \textbf{0.1905} & 9 \\
GES & 0.8182 & 0.5 & 0.4286 & 11 \\
\hline
\textbf{DMCD} & $\mathbf{0.3943 \pm 0.02}$ & $\mathbf{0.9889 \pm 0.03}$ & $0.2762 \pm 0.02$ & $\mathbf{5.9 \pm 0.54}$ \\
\bottomrule
\end{tabular}
\vspace{0.5cm}
\begin{tabular}{l l l l}
\toprule
& \textbf{Precision} $\uparrow$ & \textbf{Recall} $\uparrow$ & \textbf{F1} $\uparrow$ \\
\midrule
PC Pearson & 0.4444 & 0.5714 & 0.5000 \\
PC Spearman & 0.2857 & 0.2857 & 0.2857 \\
LiNGAM & 0.3333 & 0.2857 & 0.3077 \\
GES & 0.1818 & 0.5 & 0.2667 \\
\hline
\textbf{DMCD} & $\mathbf{0.6057 \pm 0.02}$ & $\mathbf{0.9889 \pm 0.03}$ & $\mathbf{0.751 \pm 0.02}$ \\
\bottomrule
\end{tabular}
\caption{Performance comparison on Fluxnet2015 dataset. Best scores for each metric are highlighted in \textbf{bold}.}
\label{tab:flux_results}
\end{table}

Table~\ref{tab:flux_results} compares DMCD to constraint-based (PC), functional (LiNGAM), and score-based (GES) causal discovery methods on environmental data. DMCD achieves the best performance across most metrics, including a substantial improvement in F1 score (0.751 vs. 0.5 for the nearest competitor, PC with Pearson Correlation), corresponding to an increase of approximately 50\%.

Also notable are the near-perfect true positive rate and recall, which suggest that our semantic-statistical approach is able to recover nearly all valid causal relationships in this dataset. At the same time, the false positive rate, while not the lowest among all methods, remains within a comparable range. This indicates that the exceptionally high recall is not achieved by indiscriminately predicting all possible edges, but rather by identifying exactly those that belong to the ground truth structure.

These results highlight the capabilities of DMCD on a dataset whose domain, while specialized within the earth sciences, remains relatively aligned with broad world knowledge. Compared to highly technical industrial process benchmarks like the Tennessee Eastman Process dataset, environmental variables such as temperature, radiation, and vapour pressure deficit are conceptually more accessible and more likely to be represented in general-purpose language model training data.

Next, we evaluate DMCD on IT monitoring datasets, which reflect more operational and infrastructure-related settings. This allows us to assess how well the framework generalizes to domains whose underlying relationships are less widely represented in broad-coverage textual corpora.  

\begin{table}[H]
\centering
\begin{tabular}{l l l l l l}
\toprule
 & \textbf{MoM 1} & \textbf{MoM 2} & \textbf{Ingestion} & \textbf{Web 1} & \textbf{Web 2} \\
\midrule
GCMVL & 0 & 0 & 0.20 & 0.20 & 0 \\
Dynotears & 0.26 & 0.20 & 0.14 & 0.23 & 0.30 \\
PCMCI+ & 0.40 & 0 & 0 & 0.23 & 0.30 \\
PCGCE & 0 & 0.12 & 0.12 & 0.22 & 0.15 \\
VLiNGAM & 0 & 0 & 0.19 & 0.29 & 0.18 \\
TiMINo & 0 & 0.17 & 0.18 & 0 & 0 \\
NBCB-w & 0.40 & 0 & 0.13 & 0.23 & 0.30 \\
NBCB-e & 0.13 & 0.29 & 0.27 & 0.19 & 0.42 \\
CBNB-w & 0.40 & 0 & 0.15 & 0.23 & 0.30 \\
CBNB-e & 0 & 0.24 & 0.13 & 0.22 & 0.29 \\
\hline
\textbf{DMCD} & $\mathbf{0.42 \pm 0.19}$ & $\mathbf{0.61 \pm 0.09}$ & $\mathbf{0.59 \pm 0.03}$ & $\mathbf{0.71 \pm 0.09}$ & $\mathbf{0.68 \pm 0.08}$ \\
\bottomrule
\end{tabular}
\vspace{0.4cm}
\begin{tabular}{l l l}
\toprule
 & \textbf{Antivirus 1} & \textbf{Antivirus 2} \\
\midrule
GCMVL & 0.08 & 0 \\
Dynotears & 0.18 & 0.19 \\
PCMCI+ & 0.04 & 0.11 \\
PCGCE & 0.30 & 0.45 \\
VLiNGAM & 0.15 & 0.22 \\
TiMINo & 0 & 0 \\
NBCB-w & 0.14 & 0.24 \\
NBCB-e & 0.31 & 0.45 \\
CBNB-w & 0.17 & 0.16 \\
CBNB-e & 0.31 & 0.38 \\
\hline
\textbf{DMCD} & $\mathbf{0.82 \pm 0.08}$ & $\mathbf{0.82 \pm 0.09}$ \\
\bottomrule
\end{tabular}

\caption{F1 scores comparison across IT monitoring datasets. ``1'' and ``2'' in the names denote different pre-processing strategies applied to the same underlying data. Best scores for each dataset are highlighted in \textbf{bold}.}
\label{tab:it_results}
\end{table}

Table~\ref{tab:it_results} compares DMCD against various causal discovery methods on IT monitoring datasets. DMCD achieves the highest F1 scores across all configurations, with particularly strong performance on the Antivirus datasets (0.82 compared to the next-best scores of 0.31 and 0.45) and the Web server datasets (0.71 vs. 0.29 and 0.68 vs. 0.42).

The causal structure in the IT monitoring datasets reflects specialized system architecture and workload dynamics rather than canonical scientific or common-sense relationships. The observed improvements over baselines in this setting therefore demonstrate that our proposed approach remains effective even when the domain is more operational and practice-oriented.

Across all the evaluated benchmarks, DMCD consistently achieves competitive or leading performance. These results indicate that combining metadata-informed semantic reasoning with statistical verification can offer a sizable advantage over purely data-driven methods in diverse real-world domains.

\section{Discussion}
\label{sec:discussion}

\subsection{Complexity}
\label{sec:complexity}

DMCD decomposes causal discovery into two conceptually distinct operations: LLM-based drafting and statistical verification. Its computational profile reflects this separation.

Phase I requires a single LLM inference pass over metadata describing \( n = |V|\) variables. Prompt construction scales linearly in the number of variables, $O(n)$, assuming bounded metadata per variable, and inference cost depends on the prompt and output length in tokens. Crucially, this phase does not perform explicit search over the space of possible DAGs. In contrast to classical constraint-based or score-based methods, DMCD produces a single structured hypothesis in one forward pass and does not enumerate candidate edge sets from the \( O(n^2) \) possible directed edges.

In Phase II, DMCD performs CI testing over all unordered pairs of variables in
the draft graph \( G_{\text{draft}} = (V_{\text{draft}}, E_{\text{draft}}) \),
where we denote \( k = |V_{\text{draft}}|\). Consequently, the number of CI
tests grows quadratically in \( k \): \[ C(k, 2) = \frac{k(k - 1)}{2} = O(k^2).
\] Since each test incurs a cost that depends on the number of observational
samples $N$ and the conditioning set size $|Z|$, the overall computational cost
of this phase scales as \( O( k^2 \cdot C_{\text{CI}}(N, |Z|))\).

In the worst case, when all variables in $V$ are included in the draft (i.e., \( k = n \)), this yields quadratic scaling in \( n \). In practice, however, the drafting phase typically excludes a subset of variables from $V$, namely those judged causally irrelevant, so that \( k < n \), reducing the number of tests relative to approaches that consider all variables. 

\paragraph{Comparison to existing approaches}
Classical constraint-based algorithms (e.g., PC, FCI) systematically test conditional independencies among subsets of variables, often performing repeated CI tests over the same variable pairs with progressively larger conditioning sets. Score-based methods (e.g., GES) explore the combinatorial space of possible edge sets \( E \subseteq V \times V\), which grows super-exponentially in $n$. Continuous optimization approaches (e.g., NOTEARS, GOLEM) replace discrete search with iterative global updates that typically involve repeated dense-matrix operations whose worst-case cost scales polynomially in $n$, often cubic in $n$ per iteration. Functional causal model approaches (e.g., LiNGAM variants) rely on structural assumptions and regression-based procedures whose computational cost typically scales polynomially in $n$.

DMCD does not alter the worst-case combinatorial hardness of causal structure learning. However, by generating \( G_{\text{draft}}\) from metadata prior to statistical testing, it restricts refinement to a semantically informed subset of variables and a constrained region of graph space rather than conducting a global search over all variables and edge configurations.

\subsection{Theoretical assumptions}
\label{sec:theoretical_assumptions}

DMCD builds on standard assumptions underlying graphical causal modeling and statistical independence testing.

First, we assume that the data-generating process over the observed variables can be represented by a DAG. This excludes cyclic feedback at the level of the modeled variables and renders structural features empirically informative.

Second, Phase II relies on correspondence between graph structure and conditional independence relations. In particular, it uses the principle that if the true (but unknown) causal structure underlying the data-generating process can be represented by a DAG $G^\ast$, then the observational distribution satisfies the Causal Markov condition with respect to $G^\ast$: each variable is conditionally independent of its non-descendants given its parents. This principle motivates the use of $d$-separation to derive conditional independence statements for empirical testing.

DMCD does not enforce strict satisfaction of all implied independencies, nor does it assume perfect faithfulness. Instead, discrepancies between graph-implied and data-estimated independence relations are treated as diagnostic signals for structural refinement.

Third, we assume that the conditional independence tests employed in Phase II are statistically valid under their respective modeling assumptions (e.g., linearity for partial correlation tests or sufficient sample size for Chi-squared tests). The $q$-value adjustment assumes valid $p$-values and moderate dependence among tests; in practice, these conditions may only hold approximately.

Finally, DMCD does not assume identifiability of the true causal graph from available data alone. The objective is not guaranteed recovery of $G^\ast$, but a construction of a semantically informed DAG whose structure has been empirically evaluated and refined in light of statistical evidence.

\subsection{Practical constraints}
\label{sec:practical_constraints}

While our experiments demonstrate the benefits of DMCD's approach to causal discovery, several practical factors influence where it is most effective.

First, DMCD assumes the availability of meaningful metadata. If metadata is missing, extremely terse (e.g., cryptic sensor IDs), or misleading, the advantage of the LLM-based drafting phase diminishes and the method effectively falls back to a more classical data-driven regime.
In such cases, additional human input or metadata curation would be required to fully realize the benefits of the approach.

Second, the current implementation targets problems with a moderate number of variables.
As discussed in Section~\ref{sec:complexity}, conditional independence testing remains the computational bottleneck in Phase~II, where the number of tests grows with the size of the draft graph. This reflects a general scalability challenge in causal discovery: many widely used algorithms scale at least quadratically in the number of variables, and often substantially worse, and are therefore not designed for arbitrarily large graphs without additional sparsity or structural assumptions.

Third, effective validation depends on data quality. Like other causal discovery methods that rely on statistical testing, DMCD assumes sufficiently large sample sizes, reasonable noise levels, and limited unobserved confounders. In more challenging regimes (e.g., severe missingness, heavy-tailed noise, or strong latent confounders), Phase~II may reject more candidate edges, yielding sparser graphs. In deployment settings, such outcomes should be interpreted as reflecting limitations of the data rather than shortcomings of the method itself.

\subsection{LLM data memorization risk}
\label{sec:llm_data_memorization_risk}

A potential concern when using LLMs for scientific tasks is whether performance gains arise from memorization of benchmark datasets (or, in this case, ground truth DAGs) during pre-training rather than genuine reasoning over metadata. We conducted a series of targeted probes and ablation experiments to assess this risk.

\paragraph{Sentence completion probe}
For each benchmark paper used in our evaluation, we selected a distinctive sentence and prompted the LLM used in DMCD to complete it, explicitly instructing it to draw from its knowledge of causal discovery papers. In all cases, the model hallucinated plausible but incorrect continuations. This suggests that the model does not reliably recall the original papers or their exact contents.

\paragraph{Dataset identification probe}
Next, we provided the LLM with the variable metadata for each dataset and asked it to identify the paper in which the dataset was used for causal discovery with a reported ground truth DAG. The model either responded that it could not identify such a paper or named a real but incorrect publication that did not contain a relevant DAG. This further indicates that the model does not appear to possess retrievable knowledge of the benchmark sources.

\paragraph{Ablation experiment: Tennessee Eastman Process dataset}

Our ablation experiments directly test whether DMCD relies on structural recall. The definition of the ground truth DAG for the Tennessee Eastman Process dataset in~\cite{menegozzo2022cipcad} uses neutral variable names $X_1, \dots, X_n$. We therefore removed all the other informative descriptions from the metadata and provided only these names to DMCD. Two outcomes were possible:
\begin{itemize}
    \item If DMCD still produced a graph close to the ground truth, this would suggest memorization of the DAG from pre-training.
    \item If DMCD produced a graph far from the ground truth, this would indicate that it requires variable descriptions and thus its performance depends on semantic reasoning. 
\end{itemize}

Empirically, performance worsened substantially when descriptions were removed: the F1 score dropped to $0.07 \pm 0.006$ from $0.209 \pm 0.04$, SHD increased to $146.7 \pm 51.05$ from $44.3 \pm 3.52$. This strongly suggests that DMCD does not recover the Tennessee Eastman causal structure via memorized recall, but instead relies on semantic interpretation of metadata.

\paragraph{Ablation experiment: Fluxnet2015 and IT monitoring datasets}

In the Fluxnet2015 and IT monitoring datasets, variable names are themselves informative (domain-standard abbreviations such as \emph{VPD} and \emph{NEE} in Fluxnet2015 and self-explanatory labels such as \emph{ram\_global\_prct} in the IT monitoring datasets). Thus, removing descriptions alone would not disentangle memorization from reasoning, since variable names could already encode strong semantic cues.

To address this, we renamed all variables to neutral identifiers ($X_1, \dots, X_n$), significantly rephrased their descriptions to minimize lexical overlap with the original papers while preserving semantics, and removed contextual information such as dataset names from the Phase I LLM prompt. This procedure reduces the likelihood that the model can recover a link to the source papers even if they were encountered during pre-training. 

Under this transformation, two outcomes were again possible:
\begin{itemize}
    \item If DMCD was still able to produce a DAG close to the ground truth, this would indicate successful reasoning from semantic content alone.
    \item If performance degraded severely, this would suggest dependence on lexical cues potentially tied to memorized content.
\end{itemize}

In our experiments, DMCD continued to produce high-quality DAGs (F1 score of $0.79 \pm 0.03$ for Fluxnet2015, and up to $0.88$ across IT monitoring datasets), supporting the hypothesis that it infers structure from variable semantics rather than recalling a memorized DAG. Noticeable degradation was observed only in the Middleware oriented Message activity datasets (a drop from $0.42 \pm 0.19$ to $0.33 \pm 0.13$ and from $0.61 \pm 0.09$ to $0.49 \pm 0.1$), which might indicate that the neutralized metadata in this case provides less signal for inferring stable causal structure due to removing canonical terms of this particular domain. Importantly, even with the degradation, DMCD continues to be competitive with or outperform statistical baselines. 

\paragraph{Summary}
Across probing and ablation experiments, we find no evidence that DMCD's performance is driven by memorization of benchmark DAGs. Instead, results are consistent with a mechanism in which the Phase I LLM generates structural hypotheses based on semantic interpretation of variable metadata.

\subsection{Non-determinism}
\label{sec:non_determinism}

The drafting phase of DMCD relies on an LLM, which introduces inherent non-determinism. Even with temperature set to 0.0, modern LLM APIs do not guarantee strict determinism due to implementation-level stochasticity and infrastructure-level variability. Consequently, repeated executions of DMCD on the same dataset may yield slightly different graphs.

We explicitly quantify this variability when reporting benchmarking results in Section~\ref{sec:results} by computing mean and standard deviation of metric values after 10 runs. Empirically, standard deviations are generally small, indicating that variations in draft graphs lead to only minor differences relative to the ground truth DAG. When variability is higher, it points to ambiguity in the available metadata: provided variable descriptions may not encode enough information to support a definitive causal interpretation. From this perspective, variability becomes informative: it signals lack of data rather than mere model instability.

Interestingly, this behavior parallels human expert reasoning. It has been observed, e.g., in an interactive causal discovery setting~\cite{melkas2021interactive}, that domain experts may propose slightly different causal structures depending on their background knowledge, context, or starting assumptions. This suggests that structural variation is not necessarily a flaw but a realistic reflection of knowledge-driven hypothesis generation.

Nevertheless, our future work will explore explicit stability-enhancing mechanisms for the drafting phase of DMCD, such as an LLM ``voting'' strategy in which multiple LLM runs will be executed, and only the edges proposed in the majority of outputs will be retained in the draft graph.

\section{Conclusions}
\label{sec:conclusions}

We introduced DMCD, a hybrid causal discovery framework that integrates LLM-based semantic drafting with statistical verification. Unlike traditional approaches that rely exclusively on data-driven search, DMCD leverages variable metadata to construct an informed structural prior and then validates it empirically.

Across industrial, environmental, and IT monitoring benchmarks, DMCD consistently improves recall and F1 score while maintaining competitive control of false discoveries. These results suggest that incorporating semantic reasoning into causal discovery can substantially improve structure recovery in realistic, metadata-rich settings.

Future work includes developing stability-enhancing mechanisms such as an LLM voting system for graph drafting, and investigating adaptive verification strategies in which statistical testing is guided by uncertainty in the LLM-generated structure.

While the current implementation relies on conditional independence testing, alternative validation mechanisms could also be incorporated. These include score-based model comparison using likelihood or information criteria, predictive cross-validation of structural equation models, and tests of interventional invariance when quasi-experimental data is available. More generally, any statistical criterion capable of producing structured discrepancy signals between a candidate graph and observed data could serve as input to the refinement stage.

A complementary line of future research is verification-informed adaptation of the drafting phase. In the current implementation, Phase I employs an off-the-shelf LLM without task-specific fine-tuning. However, each execution of DMCD produces an explicit statistical validation signal in Phase II that could serve as automated feedback for improving future drafts. This idea is conceptually related to reinforcement learning approaches in which LLMs are optimized through external feedback mechanisms, such as unit tests, compiler diagnostics, and execution results used as reward signals in code generation~\cite{le2022coderl,liu2023rltf, dou2024stepcoder,kulkarni2025reinforcing}. In a similar spirit, DMCD's validation layer could define structured rewards for improving generated causal hypotheses while preserving the architectural separation between semantic reasoning and statistical verification.

Finally, an open question concerns the relationship between metadata quality and causal recoverability. Systematically characterizing how semantic richness, ambiguity, and domain specificity influence structure learning performance remains an important direction for both theory and practice.

\section*{Author Contributions}

The authors are listed in alphabetical order.

Sofia Nikiforova designed and implemented DMCD, conducted the experiments, and wrote the manuscript.

Samarth KaPatel reviewed the implementation of DMCD and contributed to its refinement through code reviews and technical feedback.

Giacinto Paolo Saggese and Paul Smith provided overall conceptual guidance, contributed to architectural design decisions, and offered feedback on the manuscript.

\section*{Acknowledgements}

The authors thank Shayan Ghasemnezhad for contributions to the discussion of practical constraints and limitations, and Grigory Pomazkin for proposing the sentence completion probe used in the memorization analysis.

\bibliographystyle{unsrt}  
\bibliography{references}  

@article{ankan2024,
author = {Ankan, Ankur and Textor, Johannes},
title = {{pgmpy: a Python toolkit for Bayesian networks}},
year = {2024},
issue_date = {January 2024},
publisher = {JMLR.org},
volume = {25},
number = {1},
issn = {1532-4435},
journal = {Journal of Machine Learning Research},
month = jan,
articleno = {265},
numpages = {8}
}

@article{willig2022can,
  title={Can foundation models talk causality?},
  author={Willig, Moritz and Ze{\v{c}}evi{\'c}, Matej and Dhami, Devendra Singh and Kersting, Kristian},
  journal={arXiv preprint arXiv:2206.10591},
  year={2022}
}

@article{long2023can,
  title={Can large language models build causal graphs?},
  author={Long, Stephanie and Schuster, Tibor and Pich{\'e}, Alexandre},
  journal={arXiv preprint arXiv:2303.05279},
  year={2023}
}

@article{long2023causal,
  title={Causal discovery with language models as imperfect experts},
  author={Long, Stephanie and Pich{\'e}, Alexandre and Zantedeschi, Valentina and Schuster, Tibor and Drouin, Alexandre},
  journal={arXiv preprint arXiv:2307.02390},
  year={2023}
}

@article{jiralerspong2024efficient,
  title={Efficient causal graph discovery using large language models},
  author={Jiralerspong, Thomas and Chen, Xiaoyin and More, Yash and Shah, Vedant and Bengio, Yoshua},
  journal={arXiv preprint arXiv:2402.01207},
  year={2024}
}

@article{darvariu2024large,
  title={Large language models are effective priors for causal graph discovery},
  author={Darvariu, Victor-Alexandru and Hailes, Stephen and Musolesi, Mirco},
  journal={arXiv preprint arXiv:2405.13551},
  year={2024}
}

@inproceedings{vashishtha2025causal,
  title={{Causal order: The key to leveraging imperfect experts in causal inference}},
  author={Vashishtha, Aniket and Reddy, Abbavaram Gowtham and Kumar, Abhinav and Bachu, Saketh and Balasubramanian, Vineeth N and Sharma, Amit},
  booktitle={The Thirteenth International Conference on Learning Representations},
  year={2025}
}

@article{newsham2025large,
  title={Large language models for zero-shot inference of causal structures in biology},
  author={Newsham, Izzy and Kova{\v{c}}evi{\'c}, Luka and Moulange, Richard and Ke, Nan Rosemary and Mukherjee, Sach},
  journal={arXiv preprint arXiv:2503.04347},
  year={2025}
}

@inproceedings{li2024realtcd,
  title={{RealTCD: Temporal causal discovery from interventional data with large language model}},
  author={Li, Peiwen and Wang, Xin and Zhang, Zeyang and Meng, Yuan and Shen, Fang and Li, Yue and Wang, Jialong and Li, Yang and Zhu, Wenwu},
  booktitle={Proceedings of the 33rd ACM international conference on information and knowledge management},
  pages={4669--4677},
  year={2024}
}

@article{khatibi2024alcm,
  title={{ALCM: Autonomous LLM-augmented causal discovery framework}},
  author={Khatibi, Elahe and Abbasian, Mahyar and Yang, Zhongqi and Azimi, Iman and Rahmani, Amir M},
  journal={arXiv preprint arXiv:2405.01744},
  year={2024}
}

@article{kiciman2023causal,
  title={{Causal reasoning and large language models: Opening a new frontier for causality}},
  author={Kiciman, Emre and Ness, Robert and Sharma, Amit and Tan, Chenhao},
  journal={Transactions on Machine Learning Research},
  year={2023}
}

@article{ban2023causal,
  title={Causal structure learning supervised by large language model},
  author={Ban, Taiyu and Chen, Lyuzhou and Lyu, Derui and Wang, Xiangyu and Chen, Huanhuan},
  journal={arXiv preprint arXiv:2311.11689},
  year={2023}
}

@article{ban2025integrating,
  title={Integrating large language model for improved causal discovery},
  author={Ban, Taiyu and Chen, Lyuzhou and Lyu, Derui and Wang, Xiangyu and Zhu, Qinrui and Tu, Qiang and Chen, Huanhuan},
  journal={IEEE Transactions on Artificial Intelligence},
  year={2025},
  publisher={IEEE}
}

@article{kampani2024llm,
  title={{LLM-initialized differentiable causal discovery}},
  author={Kampani, Shiv and Hidary, David and van der Poel, Constantijn and Ganahl, Martin and Miao, Brenda},
  journal={arXiv preprint arXiv:2410.21141},
  year={2024}
}

@inproceedings{shen2025exploring,
  title={{Exploring multi-modal data with tool-augmented LLM agents for precise causal discovery}},
  author={Shen, ChengAo and Chen, Zhengzhang and Luo, Dongsheng and Xu, Dongkuan and Chen, Haifeng and Ni, Jingchao},
  booktitle={Findings of the Association for Computational Linguistics: ACL 2025},
  pages={636--660},
  year={2025}
}

@article{takayama2024integrating,
  title={{Integrating large language models in causal discovery: A statistical causal approach}},
  author={Takayama, Masayuki and Okuda, Tadahisa and Pham, Thong and Ikenoue, Tatsuyoshi and Fukuma, Shingo and Shimizu, Shohei and Sannai, Akiyoshi},
  journal={arXiv preprint arXiv:2402.01454},
  year={2024}
}

@inproceedings{ankan2023simple,
  title={A simple unified approach to testing high-dimensional conditional independences for categorical and ordinal data},
  author={Ankan, Ankur and Textor, Johannes},
  booktitle={Proceedings of the AAAI Conference on Artificial Intelligence},
  volume={37},
  number={10},
  pages={12180--12188},
  year={2023}
}

@inproceedings{melkas2021interactive,
  title={{Interactive causal structure discovery in Earth system sciences}},
  author={Melkas, Laila and Savvides, Rafael and Chandramouli, Suyog H and M{\"a}kel{\"a}, Jarmo and Nieminen, Tuomo and Mammarella, Ivan and Puolam{\"a}ki, Kai},
  booktitle={The KDD'21 Workshop on Causal Discovery},
  pages={3--25},
  year={2021},
  organization={PMLR}
}

@article{ait2023case,
  title={{Case studies of causal discovery from IT monitoring time series}},
  author={A{\"\i}t-Bachir, Ali and Assaad, Charles K and de Bignicourt, Christophe and Devijver, Emilie and Ferreira, Simon and Gaussier, Eric and Mohanna, Hosein and Zan, Lei},
  journal={arXiv preprint arXiv:2307.15678},
  year={2023}
}

@inproceedings{menegozzo2022cipcad,
  title={{CIPCaD-Bench: Continuous industrial process datasets for benchmarking causal discovery methods}},
  author={Menegozzo, Giovanni and Dall’Alba, Diego and Fiorini, Paolo},
  booktitle={2022 IEEE 18th International Conference on Automation Science and Engineering (CASE)},
  pages={2124--2131},
  year={2022},
  organization={IEEE}
}

@article{downs1993plant,
  title={A plant-wide industrial process control problem},
  author={Downs, James J and Vogel, Ernest F},
  journal={{Computers \& Chemical Engineering}},
  volume={17},
  number={3},
  pages={245--255},
  year={1993},
  publisher={Elsevier}
}

@article{reinartz2021extended,
  title={{An extended Tennessee Eastman simulation dataset for fault-detection and decision support systems}},
  author={Reinartz, Christopher and Kulahci, Murat and Ravn, Ole},
  journal={{Computers \& Chemical Engineering}},
  volume={149},
  pages={107281},
  year={2021},
  publisher={Elsevier}
}

@article{bathelt2015revision,
  title={{Revision of the Tennessee Eastman Process model}},
  author={Bathelt, Andreas and Ricker, N Lawrence and Jelali, Mohieddine},
  journal={IFAC-PapersOnLine},
  volume={48},
  number={8},
  pages={309--314},
  year={2015},
  publisher={Elsevier}
}

@article{ricker1995nonlinear,
  title={{Nonlinear modeling and state estimation for the Tennessee Eastman challenge process}},
  author={Ricker, NL and Lee, JayHyung},
  journal={{Computers \& Chemical Engineering}},
  volume={19},
  number={9},
  pages={983--1005},
  year={1995},
  publisher={Elsevier}
}

@article{chen2020complex,
  title={Complex system monitoring based on distributed least squares method},
  author={Chen, Xiaolu and Wang, Jing and Ding, Steven X},
  journal={IEEE Transactions on Automation Science and Engineering},
  volume={18},
  number={4},
  pages={1892--1900},
  year={2020},
  publisher={IEEE}
}

@article{pastorello2020fluxnet2015,
  title={{The FLUXNET2015 dataset and the ONEFlux processing pipeline for eddy covariance data}},
  author={Pastorello, Gilberto and Trotta, Carlo and Canfora, Eleonora and Chu, Housen and Christianson, Danielle and Cheah, You-Wei and Poindexter, Cristina and Chen, Jiquan and Elbashandy, Abdelrahman and Humphrey, Marty and others},
  journal={Scientific data},
  volume={7},
  number={1},
  pages={225},
  year={2020},
  publisher={Nature Publishing Group UK London}
}

@article{cheng2023causaltime,
  title={{CausalTime: Realistically generated time-series for benchmarking of causal discovery}},
  author={Cheng, Yuxiao and Wang, Ziqian and Xiao, Tingxiong and Zhong, Qin and Suo, Jinli and He, Kunlun},
  journal={arXiv preprint arXiv:2310.01753},
  year={2023}
}

@article{geffner2022deep,
  title={Deep end-to-end causal inference},
  author={Geffner, Tomas and Antoran, Javier and Foster, Adam and Gong, Wenbo and Ma, Chao and Kiciman, Emre and Sharma, Amit and Lamb, Angus and Kukla, Martin and Pawlowski, Nick and others},
  journal={arXiv preprint arXiv:2202.02195},
  year={2022}
}

@misc{munoz2020causeme,
  title={{CauseMe: An online system for benchmarking causal discovery methods}},
  author={Munoz-Mar{\'\i}, J and Mateo, G and Runge, J and Camps-Valls, G},
  note={In Preparation},
  year={2020}
}

@article{le2022coderl,
  title={{CodeRL: Mastering code generation through pretrained models and deep reinforcement learning}},
  author={Le, Hung and Wang, Yue and Gotmare, Akhilesh Deepak and Savarese, Silvio and Hoi, Steven Chu Hong},
  journal={Advances in Neural Information Processing Systems},
  volume={35},
  pages={21314--21328},
  year={2022}
}

@article{liu2023rltf,
  title={{RLTF: Reinforcement learning from unit test feedback}},
  author={Liu, Jiate and Zhu, Yiqin and Xiao, Kaiwen and Fu, Qiang and Han, Xiao and Yang, Wei and Ye, Deheng},
  journal={arXiv preprint arXiv:2307.04349},
  year={2023}
}

@article{kulkarni2025reinforcing,
  title={{Reinforcing code generation: Improving Text-to-SQL with execution-based learning}},
  author={Kulkarni, Atharv and Srikumar, Vivek},
  journal={arXiv preprint arXiv:2506.06093},
  year={2025}
}

@inproceedings{dou2024stepcoder,
  title={{StepCoder: Improving code generation with reinforcement learning from compiler feedback}},
  author={Dou, Shihan and Liu, Yan and Jia, Haoxiang and Zhou, Enyu and Xiong, Limao and Shan, Junjie and Huang, Caishuang and Wang, Xiao and Fan, Xiaoran and Xi, Zhiheng and others},
  booktitle={Proceedings of the 62nd Annual Meeting of the Association for Computational Linguistics (Volume 1: Long Papers)},
  pages={4571--4585},
  year={2024}
}

@article{storey2003statistical,
  title={Statistical significance for genomewide studies},
  author={Storey, John D and Tibshirani, Robert},
  journal={Proceedings of the National Academy of Sciences},
  volume={100},
  number={16},
  pages={9440--9445},
  year={2003},
  publisher={National Academy of Sciences}
}

@article{ma2025causal,
  title={{Causal inference with large language model: A survey}},
  author={Ma, Jing},
  journal={Findings of the Association for Computational Linguistics: NAACL 2025},
  pages={5886--5898},
  year={2025}
}

@inproceedings{wan2025large,
  title={{Large language models for causal discovery: Current landscape and future directions}},
  author={Wan, Guangya and Lu, Yunsheng and Wu, Yuqi and Hu, Mengxuan and Li, Sheng},
  booktitle={Proceedings of the Thirty-Fourth International Joint Conference on Artificial Intelligence},
  pages={10687--10695},
  year={2025}
}

@techreport{causify2026datamap,
  title        = {{Causify DataMap: Automatic causal probabilistic reasoning}},
  institution  = {{Causify AI}},
  year         = {2026},
  type         = {Internal Technical Report},
}

\end{document}